\documentclass[letterpaper, 10pt, conference]{ieeeconf}

\IEEEoverridecommandlockouts
\usepackage{algorithm,algorithmic}
\usepackage{graphicx}
\usepackage{amsmath}
\usepackage{amssymb}
\allowdisplaybreaks

\usepackage{cite}

\newcommand{\vb}[1]{\boldsymbol{#1}}

\usepackage[final, defaultcolor=red]{changes}

\begin{document}
\bstctlcite{IEEEexample:BSTcontrol} 

\title{Proprioception and Tail Control Enable Extreme\\ Terrain Traversal by Quadruped Robots}

\author{Yanhao Yang\textsuperscript{1, 2}, Joseph Norby\textsuperscript{1}, Justin K. Yim\textsuperscript{1, 3}, and Aaron M. Johnson\textsuperscript{1}
\thanks{This work is an extended version of the workshop abstracts \cite{yang2022proprioception,yang2021improving}.}
\thanks{This research was sponsored by the Army Research Office under Grant Number W911NF-19-1-0080 and the National Science Foundation under Grant Number CCF-2030859 to the Computing Research Association for the CIFellows Project. The views and conclusions contained in this document are those of the authors and should not be interpreted as representing the official policies, either expressed or implied, of the Army Research Office, the National Science Foundation, or the U.S. Government. The U.S. Government is authorized to reproduce and distribute reprints for Government purposes notwithstanding any copyright notation herein.}
\thanks{\textsuperscript{1}Department of Mechanical Engineering, Carnegie Mellon University, Pittsburgh, PA 15213, USA, {\tt\small \{jnorby, amj1\}@andrew.cmu.edu}}
\thanks{\textsuperscript{2}Collaborative Robotics and Intelligent Systems Institute, Oregon State University, Corvallis, OR 97331 USA, {\tt\small yangyanh@oregonstate.edu}}
\thanks{\textsuperscript{3}Department of Mechanical Science and Engineering, University of Illinois Urbana-Champaign, Urbana, IL 61801 USA {\tt\small jkyim@illinois.edu}}
}

\maketitle

\begin{abstract}

Legged robots leverage ground contacts and the reaction forces they provide to achieve agile locomotion. However, uncertainty coupled with contact discontinuities can lead to failure, especially in real-world environments with unexpected height variations such as rocky hills or curbs. To enable dynamic traversal of extreme terrain, this work introduces 1) a proprioception-based gait planner for estimating unknown hybrid events due to elevation changes and responding by modifying contact schedules and planned footholds online, and 2) a two-degree-of-freedom tail for improving contact-independent control and a corresponding decoupled control scheme for better versatility and efficiency. Simulation results show that the gait planner significantly improves stability under unforeseen terrain height changes compared to methods that assume fixed contact schedules and footholds. Further, \added{tests have shown that the tail is particularly} effective at maintaining stability when encountering a terrain change with an initial angular disturbance. The results show that these approaches work synergistically to stabilize locomotion with elevation changes up to 1.5 times the leg length and tilted initial states.

\end{abstract}

\section{Introduction}

\begin{figure}[t]
    \centering
    \includegraphics[width=\linewidth]{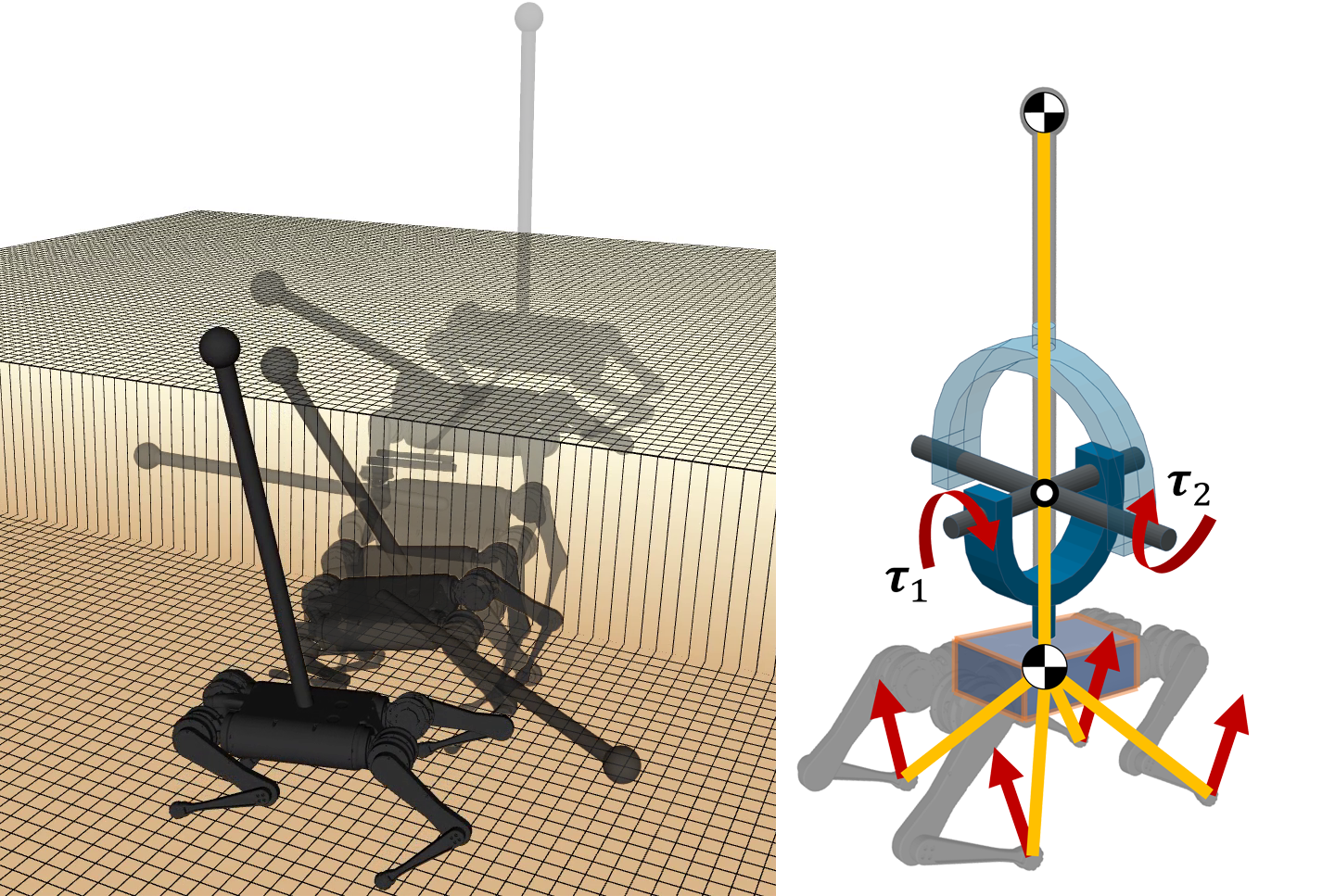}
    \caption{Left: A quadruped robot successfully traverses an unforeseen cliff using animal-inspired proprioception and tail. \added{Right: Schematic diagram of the control modeling of the tailed robot.}}
    \label{fig: motivation}
\end{figure}

\looseness=-1Quadrupedal robots have great potential because of their ability to traverse challenging environments that wheeled and tracked robots cannot. This makes them ideal for tasks such as environmental monitoring and disaster relief. But performing these tasks requires the ability to safely traverse extremely uneven terrains, such as rocky hills or curbs (\added{the left panel of} Fig.~\ref{fig: motivation}). While legged locomotion controllers can be robust, most of them rely on predefined nominal contact schedules or heuristic-based nominal footholds \cite{raibert1986legged, pratt2006capture, di2018dynamic}. These methods are good at walking in relatively flat laboratory environments, but cannot easily handle extreme terrains, where elevation changes are large enough to invalidate nominal contact schedules and footholds. In contrast, animals can easily traverse these environments using various strategies, including placing feet in repeated locations to ensure reliable contact \cite{christensen2004outwitting}, using tails to reject disturbances \cite{libby2012tail,shield2021tails}, and distributed limb control to promote rapid and reactive behaviors \cite{ijspeert2008central}. These biological phenomena inspire us to propose new approaches for the perception and control of quadruped robots to improve robustness across extreme terrains.

\looseness=-1One common approach to improve performance on rough terrain is to incorporate perception. Exteroceptive sensors such as depth cameras or lidars are common for legged robots, providing an almost out-of-the-box mapping of the environment. 
Based on terrain maps constructed from perception, we can use online planning with foothold search algorithms \cite{jenelten2020perceptive}, or \deleted{even run} real-time controllers like hybrid iLQR \cite{kong2022hybrid}. However, these algorithms rely on known and accurate terrain maps. In unknown real-world environments, terrain maps constructed online may not be perfect due to sensor noise, obstacles, reflections, or lighting conditions \cite{miki2022learning}. Proprioception is another option for adapting to unstructured terrain and has proven to be powerful and trustworthy. Proprioception on quadruped robots can be achieved by \added{contact sensing \cite{hutter2016anymal, lysakowski2022unsupervised}} or generalized disturbance observers that \added{utilize} highly transparent actuation \cite{johnson2010disturbance,wensing2017proprioceptive,kenneally2018actuator}. These can then be combined with probabilistic terrain estimation \cite{fankhauser2018probabilistic}, event-based control \cite{bledt2018contact}, \added{machine learning \cite{lysakowski2022unsupervised, lee2020learning, miki2022learning}}, and exteroceptive information \cite{homberger2019support, miki2022learning} to improve robot robustness in unknown terrain. Although these approaches are promising, they are still limited by the kinematics of the leg joints to traverse extreme environments with large height changes.

Another approach is to design \added{robots} with dynamics that are stable with respect to contact errors, including passive stabilization and \added{contact-independent actuation}. Swing-leg retraction \cite{seyfarth2003swing}\deleted{, also observed in biology,} can passively improve gait stability on uneven terrain by changing the shape of the hybrid guard \cite{zhu2022hybrid}. However, it cannot actively respond to extreme height changes and has a basin of attraction limited by the stroke of the leg. \added{Instead, some researchers have implemented contact-independent actuation methods for stabilizing robots. These methods involve utilizing existing limbs \cite{kurtz2022mini} or incorporating additional actuators like tails \cite{libby2016comparative} or reaction wheels \cite{lee2022reaction}.} In \cite{briggs2012tails, fawcett2021real}, a quadruped was equipped with a tail and used it to effectively suppress impulsive perturbations, and in \cite{norby2021enabling}, the authors proposed an aerodynamic tail that utilizes drag to improve efficiency. However, given the limited angular deflection, tails may require nonholonomic behavior like conic motion to maximize maneuverability \cite{patel2015conical}. In addition, the tail motion also needs to be coordinated with the legs to prevent instability caused by mutual interference.

\begin{figure}[t]
    \centering
    \includegraphics[width=.9\linewidth]{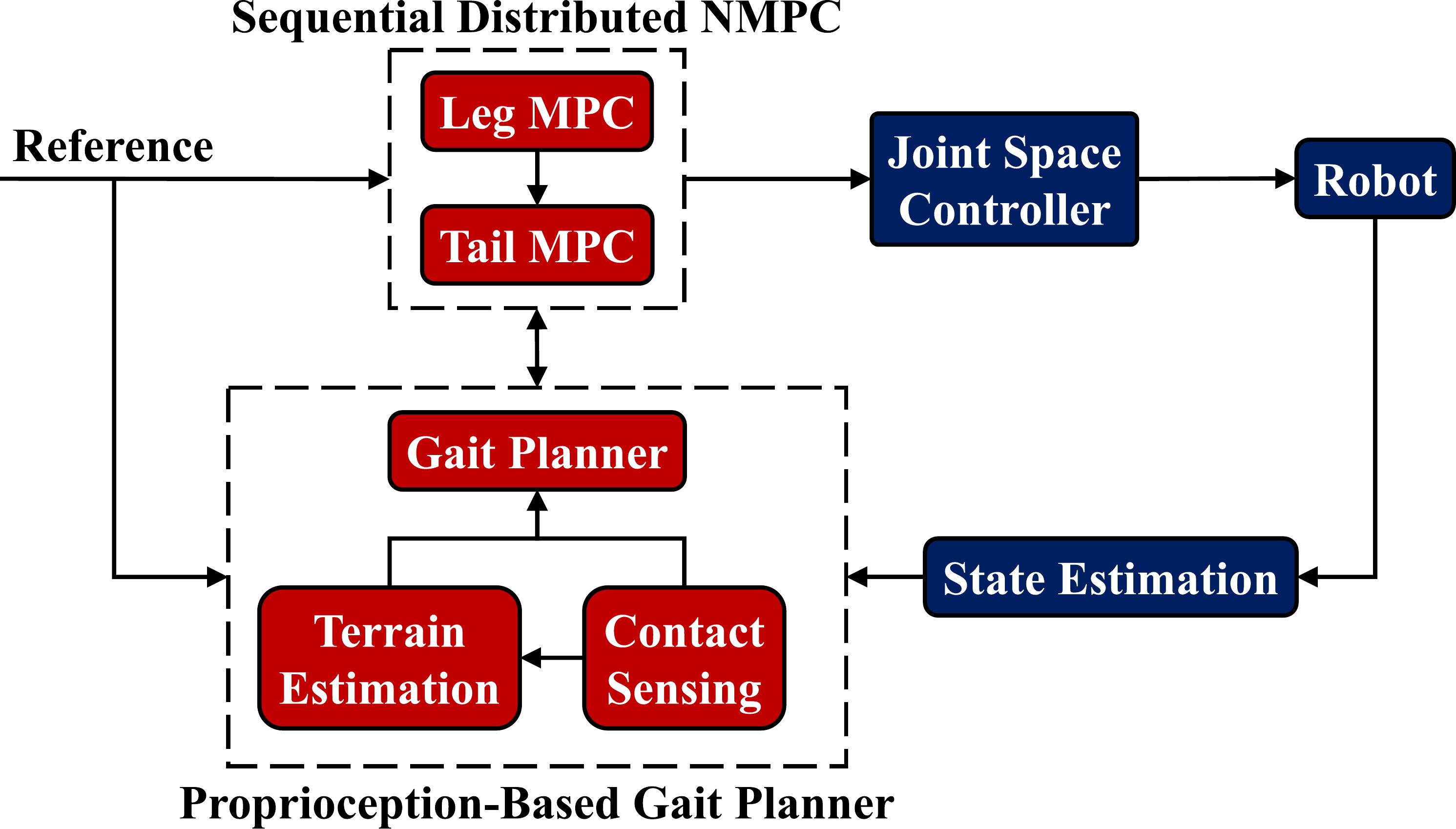}
    \caption{Block diagram of the proposed control scheme. The system consists of three main parts: proprioception-based gait planner (bottom left red blocks), sequential distributed NMPC for leg and tail control (top left red blocks), and robot interface (blue blocks on right). Among them, the red blocks are the main contributions of this paper.}
    \label{fig: block_diagram}
\end{figure}

This work proposes a strategy that incorporates perception, control, and tails to synergistically improve extreme terrain traversal. Specifically, as illustrated in Fig.~\ref{fig: block_diagram}, we propose: 1) a gait planner that online updates contact, body, foothold, and terrain references based on \added{proprioception} to \added{adapt} unexpected hybrid events and elevation changes (Sec.~\ref{sec: proprioception_based_gait_planner}), and 2) a time-scale decoupled sequential distributed nonlinear model predictive controller (NMPC) that enables high tail maneuverability within the limited joint angles of a 2-degree-of-freedom (2DOF) tail on a quadruped robot (Sec.~\ref{sec: tailed_quadruped_robot_control}). 

We conduct four simulation experiments (Sec.~\ref{sec: experimental_evaluation}) to evaluate the proposed perception and control algorithms on challenging terrains, both \added{individually} and in combination. We first measure the accuracy of the proprioceptive terrain estimation by walking a quadruped robot on unknown rough terrain and computing the estimation error (Sec.~\ref{sec:terrainresults}). We then measure the performance of the proposed sequential distributed NMPC for leg and tail control and compare it against \deleted{other control methods --} feedback control and centralized \added{NMPC} \deleted{--} in an environment where the robot misses a contact and needs to react quickly to \added{maintain} balance (Sec.~\ref{sec:nmpc_comparison}). We lastly evaluate the effect of these methods on extreme terrain traversal by simulating the robot walking down large, unexpected steps. We show that the proprioception-based gait planner improves locomotion stability (Sec.~\ref{sec:proprioresults}), and that coordinating 2-DOF tail motion with the legs further improves stability in the presence of an angular disturbance (Sec.~\ref{sec:tailresults}). Finally, in Sec.~\ref{sec: conclusion_and_future_work}, we discuss conclusions and future work.

\section{Proprioception-based Gait Planner}
\label{sec: proprioception_based_gait_planner}

The proprioception-based gait planner is designed to identify unexpected hybrid events online, estimate unknown terrain, and adapt to unforeseen elevation changes by modifying the gait including contact schedules, planned footholds, and reference body and swing foot trajectories. It is divided into two steps: proprioception and gait planning. The goal of proprioception is to perform contact sensing and generate terrain estimates. Based on this information, the gait planner refines the \deleted{nominal} contact schedules and reference trajectories, allowing adaptation to unknown terrain, and repositions the foot for more robust support.

\subsection{Contact Sensing Finite-State Machine}
\label{sec: contact_sensing_finite_state_machine}

Legged robots as hybrid systems rely heavily on correct contact information to control and plan ground reaction forces (GRF) to traverse unstructured terrain. In this section, we discuss the implementation of a proprioception-based finite-state machine and use it to perform contact sensing.

\begin{figure}[t]
    \centering
    \includegraphics[width=.7\linewidth]{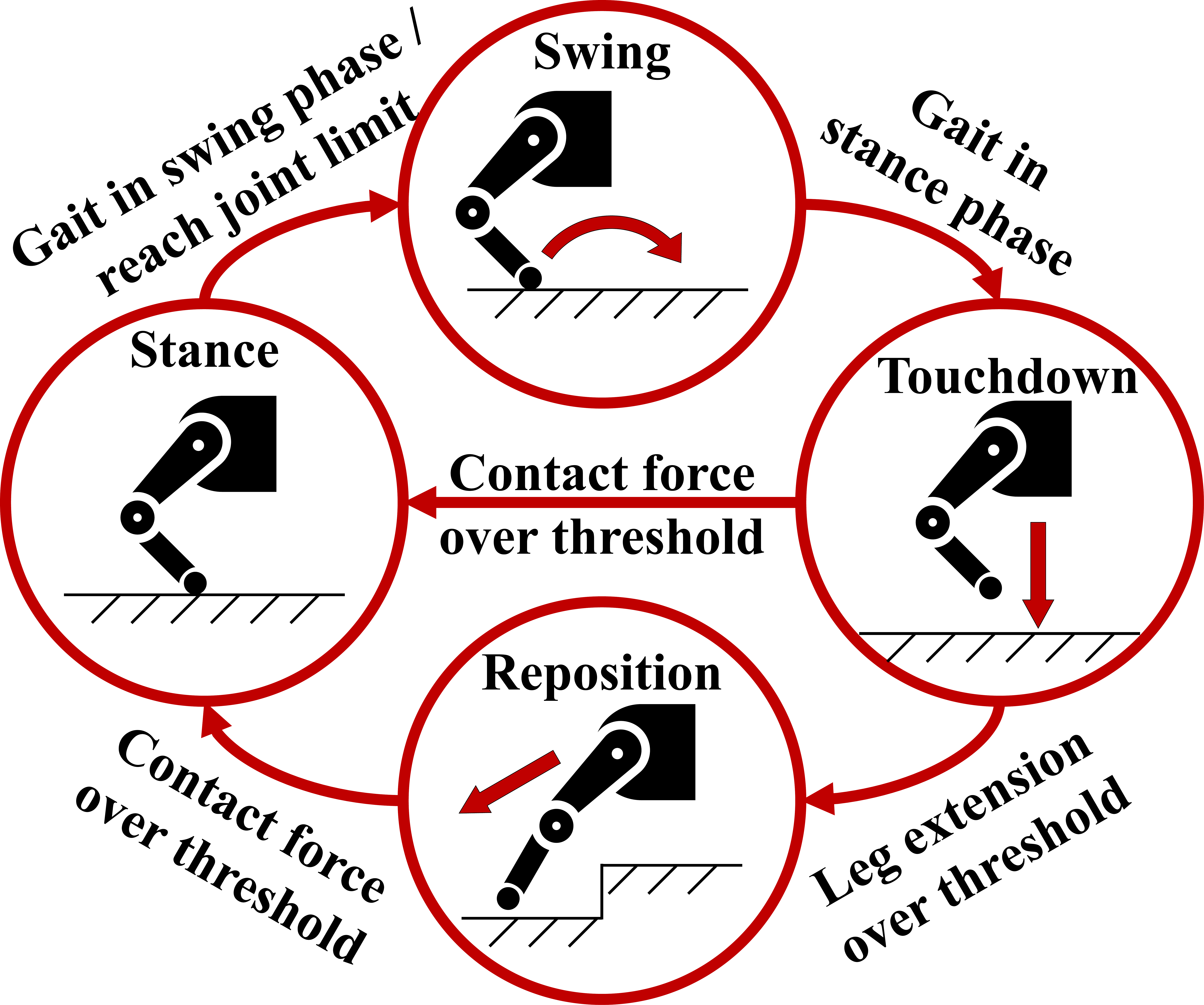}
    \caption{Illustration of the contact sensing finite state machine. The finite state machine detects the contact state of each leg of the robot according to the listed conditions. It notifies the robot if the contacts are running on a predetermined schedule or experiencing an unexpected loss of contact.}
    \label{fig: contact_sensing_finite_state_machine}
\end{figure}

As illustrated in Fig.~\ref{fig: contact_sensing_finite_state_machine}, \added{we subdivide the clock-based nominal stance-swing biphasic contact schedule into 4 distinct stages: stance, swing, touchdown, and reposition. Each stage is accompanied by a corresponding duration and switching conditions including time, leg extension, and contact force.} When the clock is about to transition to the nominal stance phase, the legs begin to touchdown. By monitoring leg extension during the touchdown \added{stage}, a missed contact can be identified. In that case, the finite-state machine will claim a contact loss and switch to the repositioning \added{stage} to avoid hyperextension, see Section~\ref{sec: gait_planner} for details. On the other hand, if solid support can be confirmed by examining the contact force obtained by a momentum observer or contact sensor, the finite-state machine triggers the transition to the stance \added{stage}. Finally, early liftoffs can be performed when the leg reaches kinematic limits to avoid physical hard stops. Note that, unlike \cite{bledt2018contact}, we only focus on the case of delayed contact, because it breaks nominal support. The effect of early contact can be minimized by force control that applies a low force until the intended contact time.

\subsection{Terrain Estimation}
\label{sec: terrain_estimation}

Successful planning and control of legged robots on rough terrain requires not only contact information but also a good understanding of the height variation of the terrain. Here we propose a terrain estimation algorithm that aims to estimate unknown elevation changes by filtering the foothold history based on contact sensing information. It provides terrain estimation updates for both swing and stance legs, and is designed to use limited sensors to build maps accurate enough to aid future control and planning.

\begin{algorithm}[t]
\caption{Terrain Estimation}
\begin{algorithmic}[1]
\label{alg: terrain_estimation}
\renewcommand{\algorithmicrequire}{\textbf{Input:}}
\renewcommand{\algorithmicensure}{\textbf{Output:}}
\REQUIRE terrainEst, swingSpace, contactState, footPos
\ENSURE  terrainEst, swingSpace
\FOR {legIdx $= 0$ to $3$}
\IF {!contactState.at(legIdx)}
\IF {footPos.z $<$ terrainEst.at(footPos.xy)}
\STATE terrainEst.at(footPos.xy) $\gets$ footPos.z\;
\STATE swingSpace.at(legIdx).append(footPos.xy)\;
\ENDIF
\ELSE
\STATE terrainEst.at(footPos.xy) $\gets$ footPos.z\;
\WHILE {!swingSpace.at(legIdx).empty()}
\STATE terrainEst.at(swingSpace.at(legIdx).pop\_back()) $\gets$ footPos.z\;
\ENDWHILE
\ENDIF
\ENDFOR
\end{algorithmic} 
\end{algorithm}

\looseness=-1The terrain estimation process is summarized in two steps in Algorithm~\ref{alg: terrain_estimation}. For initialization, \texttt{terrainEst} is a 2D grid storing elevation information and \texttt{swingSpace} is a list temporarily storing swing foot positions in the grid. First, the algorithm checks the contact sensing, \texttt{contactState}. For a non-contact foot, including swing or missed contact, we determine if it is lower than the terrain estimate \texttt{terrainEst} at the current position \texttt{footPos.xy}. If so, the current terrain estimate for this location is too high, but the actual altitude is uncertain. The algorithm temporarily records the current foot height \texttt{footPos.z} so that the reference trajectory can be updated to notify the NMPC of a terrain change and allow it to lower the body height. But since the foot has not touched the ground at this point, the actual ground height is lower than this temporary value. So we also save that position in \texttt{swingSpace} and wait for the next touchdown to update it. Second, if it is in contact, we save the current foot position in the terrain estimation history and update all temporary values in \texttt{swingSpace} with it. This assumes a simple vertical drop on the terrain where the foot misses contact so that it can better plan the swing trajectory of the rear foot to make a direct touchdown, rather than missing contact again. Finally, we post-process the terrain estimate \cite{fankhauser2016universal}, including inpainting unfilled regions according to their neighborhoods and applying a convolutional averaging filter for smoothing. Note that, as in \cite{homberger2019support}, proprioception-based terrain estimation can also be extended to incorporate exteroceptive sensing via an adaptive weighting scheme.

\subsection{Gait Planner}
\label{sec: gait_planner}

Given contact sensing information and the estimated terrain, it is important to plan an appropriate gait, including contact schedules and foot trajectories, to traverse this environment stably. In this section, we propose a gait planner that independently plans contact and foothold sequences for each leg to reflect the terrain and converges asymptotically to the central clock-based gait.

\begin{figure}[t]
    \centering
    \includegraphics[width=\linewidth]{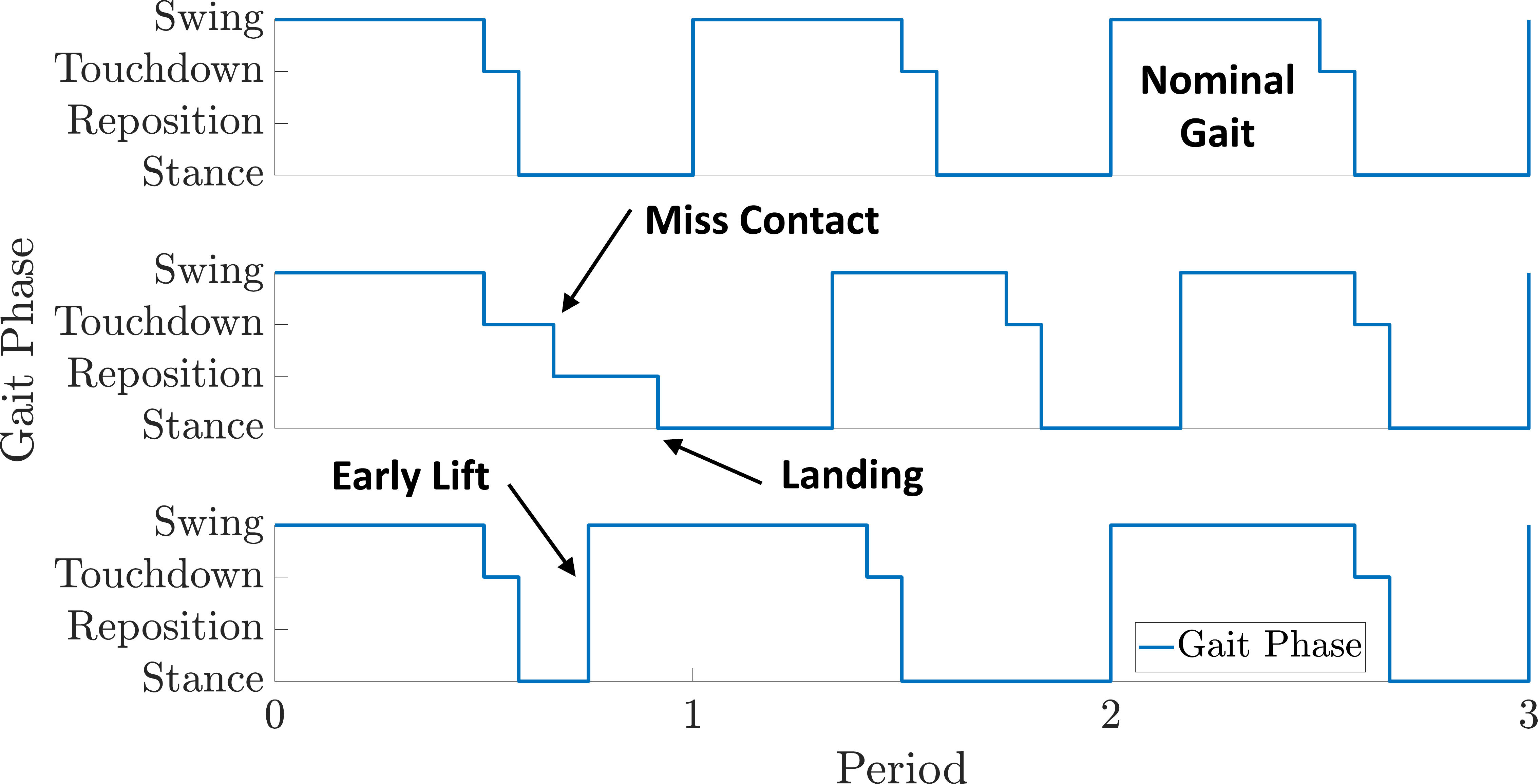}
    \caption{Illustration of gait planning for the robot during falling and landing. The gait plan is based on three discrete events in the contact schedule: 1) miss contact, 2) landing, and 3) early lift. Top: nominal clock-based gait. Middle: gait plan responds to missed contact and landing, then converges back to the clock-based gait. Bottom: gait plan lifts the trapped leg early and \added{then converges back to the clock-based gait.}}
    \label{fig: gait_planner}
\end{figure}

In order to reduce the impact on NMPC warm-starting, the gait planner is designed to modify the contact schedule as little as possible. We define three discrete events in the contact schedule: 1) miss contact, 2) landing, and 3) early lift. Fig.~\ref{fig: gait_planner} shows the process for recovery to the nominal gait when the robot misses contact or lifts off early. Specifically, for missed contacts, we assume that the entire current stance \added{stage} is lost until the landing is triggered, to guarantee sufficiently conservative behaviors. When the landing is confirmed, we perform a full stance \added{stage} to ensure enough time to absorb the kinetic energy of the fall. Finally, for early lift due to kinematic constraints, we assign a full swing \added{stage} to make sure the leg is able to clear any obstacles. In this way, once the event occurs, we can immediately determine the entire contact schedule during the adaptive gait.

For events occurring at $t_{\text{event}}$, we switch to an adaptive gait with an offset phase of the reference time $t_{\text{adaptive}}$, and converge linearly to the original gait during the event period $T_{\text{event}}$. In this paper, $T_{\text{event}}$ is set as $T_{\text{event}} = 2T_{\text{nominal}}$ to ensure that the robot returns to the nominal gait in time. Specifically, given a nominal gait period $T_{\text{nominal}}$, the applied gait phase $\phi(t)$ is the weighted average of the nominal and adaptive gait phases at reference times $t_{\text{nominal}}$ and $t_{\text{adaptive}}$, respectively. The weight $w(t)$ varies linearly according to the event time $t_{\text{event}}$ and the event period $T_{\text{event}}$,
\begin{subequations}
\begin{align}
\phi(t)&= (1-w)\phi_{\text{nominal}} + w\phi_{\text{adaptive}} \\
&= (1-w) \frac{\bmod \left(t-t_{\text {nominal}}, T_{\text {nominal}}\right)}{T_{\text {nominal}}} \nonumber \\
&\quad+w \frac{\bmod \left(t-t_{\text {adaptive}}, T_{\text {nominal}}\right)}{T_{\text {nominal}}} \\
w(t)&=\max \left(1-\frac{t-t_{\text {event}}}{T_{\text {event}}}, 0\right)
\end{align}
\end{subequations}

Similar to Raibert's heuristic \cite{raibert1986legged, pratt2006capture}, each desired footstep point $\vb{p}_l$ is determined from the NMPC-predicted trajectories as well as dynamics and kinematics heuristics as
\begin{equation}
 \vb{p}_{l}=\begin{cases} \vb{p}_{\text {center}}+\vb{p}_{\text {vel}}+\vb{p}_{\text {centrifugal}} & \text{(stance)} \\ \vb{R}_{v}\vb{p}_{0} & \text{(reposition)} \end{cases}
\end{equation}
where $\vb{p}_{0}$ is the nominal foot position when the leg is stretched downwards and $\vb{R}_{v}$ is the rotation matrix that rotates the nominal leg extension for the repositioned foothold, which is designed to maximize the possible supporting polygons when landing. The current method simply rotates the nominally downwards leg towards the velocity direction to the joint limit. An optional approach is to parameterize the desired location by referring to the capture point, as discussed in \cite{pratt2006capture}. $\vb{p}_{\text {center}}$ formulates a minimum enclosing circle problem for the stance center to minimize the maximum distance from the leg bases $\vb{p}_{\text {lb}}$ during the stance phase, which ensures foothold reachability,
\begin{align} \vb{p}_{\text {center}} &=\underset{\vb{p}}{\operatorname{argmin}}\left(\underset{i \in \text{stance}}{\max}\left\|\vb{p}-\vb{p}_{lb, i}\right\|_{2}^{2}\right) \end{align}
\looseness=-1Offset terms $\vb{p}_{\text {vel}}$ and $\vb{p}_{\text {centrifugal}}$ based on velocity and angular velocity tracking \cite{pratt2006capture} are added to the nominal foot position to minimize undesired moments \deleted{caused by GRFs} during agile motion,
\begin{subequations}
\begin{align} \vb{p}_{\text {vel}} &=\sqrt{\frac{p_{lb_{z}, \text{td}}}{g}}\left(\dot{{\vb{p}}}_{b, \text{td,ref}}-\dot{\vb{p}}_{b, \text{td}}\right) \\ \vb{p}_{\text {centrifugal}} &=\frac{p_{lb_{z}, \text{td}}}{g} \dot{\vb{p}}_{b, \text{td}} \times \vb{\omega}_{\text{ref}}
\end{align}
\end{subequations}
where $\vb{p}_{b}$ is the body position, the subscript \texttt{td} and \texttt{ref} denotes the value at touchdown and in the reference trajectory, and $g$ is the gravity acceleration. The swing foot trajectory is then generated from discrete footholds and swing apexes using cubic Hermite splines. The swing apex is determined by the nominal hip height and is shifted to be between the predicted body height and the estimated terrain height.

\section{Tailed Quadruped Robot Control}
\label{sec: tailed_quadruped_robot_control}

To complement the kinematically constrained legs under hybrid perturbations, we design a tailed quadruped robot controller to use a 2DOF tail to improve extreme terrain traversal performance. The controller accounts for the nonlinear dynamics of the tail and enables nonholonomic tail behavior in SE(3). Furthermore, it is compatible with existing leg controllers and operates asynchronously.

\subsection{Tailed Robot Centroidal Dynamics}
\label{sec: tailed_robot_centroidal_dynamics}

\added{The nonlinear dynamics of tailed robots can be simplified to a system consisting of a single rigid body and a point-mass tail, with GRF inputs and internal tail torques, as shown in the right panel of Fig.~\ref{fig: motivation}}. Define the system state space as
\begin{equation}
 \vb{x}\added{=\left[\begin{array}{ll}\vb{q} & \vb{v}\end{array}\right]^{\top}}=\left[\begin{array}{llllll}\vb{p}_{b} & \vb{\theta} & \vb{\phi} & \dot{\vb{p}}_{b} & \vb{\omega} & \dot{\vb{\phi}}\end{array}\right]^{\top} 
\end{equation}
where $\vb{p}_{b}$ is the position of the robot in the world frame, $\vb{\theta}$ is the Euler angles describing the body orientation in ZYX (an alternative approach that avoids singularities in the Euler angles is the representation-free model \cite{ding2021representation}), $\vb{\phi}$ is the position of tail \added{joints}, and $\vb{\omega}$ is the robot angular velocity in the body frame. The system input can be divided into leg GRFs $\vb{u}_l$ and tail \added{joint} torques $\vb{u}_t$,
\begin{equation}
 \vb{u}=\left[\begin{array}{ll}\vb{u}_{l} & \vb{u}_{t}\end{array}\right]^{\top} = \left[\begin{array}{llllll}\vb{f}_{1} & \vb{f}_{2} & \vb{f}_{3} & \vb{f}_{4} & \tau_{1} & \tau_{2}\end{array}\right]^{\top}
\end{equation}
The state-space dynamics model is derived \added{by applying Lagrange's equations}
\begin{align}
 L &\added{= \frac{1}{2}\left(\vb{v}^{\top}\vb{J}_{v, b}^{\top}\vb{M}_{b}\vb{J}_{v, b}\vb{v} + \vb{v}^{\top}\vb{J}_{v, t}^{\top}\vb{M}_{t}\vb{J}_{v, t}\vb{v}\right)} \nonumber\\
 &\quad-\added{\left(m_{b}p_{b_{z}} + m_{t}p_{t_{z}}\right)g}
\end{align}
\begin{equation}
 \frac{d}{dt}\left(\frac{\partial L}{\partial \vb{v}}\frac{\partial \vb{v}}{\dot{\vb{q}}}\right) - \frac{\partial L}{\vb{q}} = \left[\begin{array}{ll} \vb{J}_{u, b} & \vb{J}_{u, t}\end{array}\right]\added{\vb{u}}
\end{equation}
\added{where $\vb{J}_{v, b}$ and $\vb{J}_{v, t}$ correspond to the Jacobians that map the state-space velocities to the body velocities of the robot's body and tail, while $\vb{J}_{u, t}$, and $\vb{J}_{u, t}$ represent the Jaocbians that map the control input to the generalized coordinates. $\vb{M}_{b}$, $\vb{M}_{t}$, $m_{b}$, and $m_{t}$ correspond to the inertia matrix and mass of the body and tail. $p_{b_{z}}$ and $p_{t_{z}}$ denote the heights of the body and tail relative to the world, which can be directly obtained from the state or calculated using forward kinematics. The continuous dynamics is subsequently discretized using the implicit form of the backward Euler method for time step $t_i$,  resulting in the equation $f_d(\vb{x}_i, \vb{x}_{i+1}, \vb{u}_i, \vb{p}_{l, i}, t_i)=\vb{0}$}, where $\vb{p}_{l, i}$ is the foot position.


\subsection{Time-Scale Decoupled Sequential Distributed Model Predictive Control}
\label{sec: time_scale_decoupled_sequential_distributed_model_predictive_control}

\looseness=-1It is not desirable to force the legs and tail to use the same controller for multiple reasons \cite{yang2021improving}. First, many existing leg controllers have achieved remarkable success \cite{raibert1986legged, di2018dynamic}, \added{so one of our goals is to} be able to reuse these successful controllers. Second, including the legs and tail in the same controller simultaneously increases the dimensionality of the system and the complexity of multibody dynamics, thus impairing robustness. In addition, \cite{heim2016designing} has shown that, with appropriate approximations, tail dynamics can be decoupled from states other than body pose. Therefore, it is reasonable to decouple the control of tail and legs. Previous work designed distributed NMPC frameworks for such partially decoupled systems and discussed their closed-loop stability \cite{liu2010sequential}. \added{Another prior study applied a similar concept to develop a distributed quadratic programming-based controller for quadruped robots and explored its periodic orbital stability and asymptotic convergence \cite{kamidi2022distributed}. Inspired by these prior studies, this section introduces a tail controller based on distributed NMPC to decouple the control of tail and legs.} While this choice will lead to some suboptimality, the ability to reuse existing leg controllers and run each NMPC at a faster rate makes it a desirable approach (as we show in Sec.~\ref{sec:nmpc_comparison}).

We design a sequential distributed NMPC based decoupled control for leg and tail, summarized in the top red blocks in Fig.~\ref{fig: block_diagram}, which allows the tail to account for the leg-driven effect without specifying a particular form of leg controller. Given the scheduled footstep position, the leg controller solves the GRFs based on the centroidal dynamics NMPC. This information is passed to the tail controller, which solves the tail torque. Note that this leg controller can be replaced with any existing controller as long as the approximate net torque on the body from the GRFs is known. The required leg information can be GRFs, contact sequence, or joint torque commands. In this way, we can reuse the existing leg controller, reduce the information required for the tail, and run the leg and tail NMPC in parallel.

Specifically, in this work, we use the \added{leg} NMPC from \cite{norby2022quad} to determine the GRFs of leg control and then send the solution to \added{the} tail NMPC to solve for the tail torque. The tail NMPC maintains the nonlinearity of SE(3) dynamics. \deleted{The hybrid system is simplified to a switched system with the contact schedule from the gait planner.} Given the desired body state trajectory $\vb{x}_{i, \text{ref}}$, nominal input $\vb{u}_{i, \text{ref}}$, planned foot position $\vb{p}_{l, i}$, initial condition $\vb{x}_{\text{initial}}$, and known leg control GRFs $\vb{u}_{l, \text{sol}}$, the tail NMPC can be formulated as the following optimization problem,
\begin{subequations}\label{eq:mpc}
\begin{align}
  \min _{\vb{x}, \vb{u}} \quad & \sum_{i=0}^{N-1}\left\|\vb{x}_{i+1}-\vb{x}_{i+1, \text{ref}}\right\|_{\vb{Q}_{i}}+\left\|\vb{u}_{i}-\vb{u}_{i, \text{ref}}\right\|_{\vb{R}_{i}} \\ 
  s.t. \quad & \vb{x}_{0}=\vb{x}_{\text{initial}} \quad \text{(initial condition)} \\ 
  & f_{d}\left(\vb{x}_{i}, \vb{x}_{i+1}, \vb{u}_{i}, \vb{p}_{l, i}, t_{i}\right)=\vb{0} \quad \text{(dynamics)}\\ 
  & \vb{x}_{i} \in \mathbb{X} \quad \text{(state bound)} \\ 
  & \vb{u}_{i} \in \mathbb{U} \quad \text{(control bound)} \\ 
  & \vb{u}_{l} = \vb{u}_{l, \text{sol}} \quad \text{(known GRF)} \\ 
  & \vb{C}_{i} \vb{u}_{i} \leqslant \vb{0} \quad \text{(friction pyramid)} \\
  & \vb{D}_{i} \vb{u}_{i}=\vb{0} \quad \text{(contact selection)}
\end{align}
\end{subequations}
where $i\in [0, \ldots, N-1]$ is the time index, $N$ is the horizon length, $\vb{Q}_i$ and $\vb{R}_i$ are the diagonal quadratic cost matrix for state and input, $t_{i}$ is the finite element duration, $\mathbb{X}$ and $\mathbb{U}$ are the feasible state and control set, $\vb{C}_i$ is the friction pyramid matrix, \added{and the hybrid system is simplified through the contact selection matrix $\vb{D}_i$}.

We also propose a novel warmstart technique to decouple the time scale between the NMPC update rate and finite element discretization, allowing asynchronous solving of leg and tail NMPCs. Specifically, we make the duration of the first finite element \added{$t_{0}$} adjustable to allow the subsequent ones to remain aligned with the discretized collocation points. In this way, we can start solving the NMPC at any time. This allows rapid reaction to the latest state estimates and proprioception updates, and the tail and leg NMPC to run asynchronously (while still passing control information to each other). In addition, since the NMPC problem can be solved multiple times between the two collocation points and the rest remain unshifted, this greatly reduces modeling error accumulation and improves warmstarting.

\begin{figure}[t]
    \centering
    \includegraphics[width=.9\linewidth]{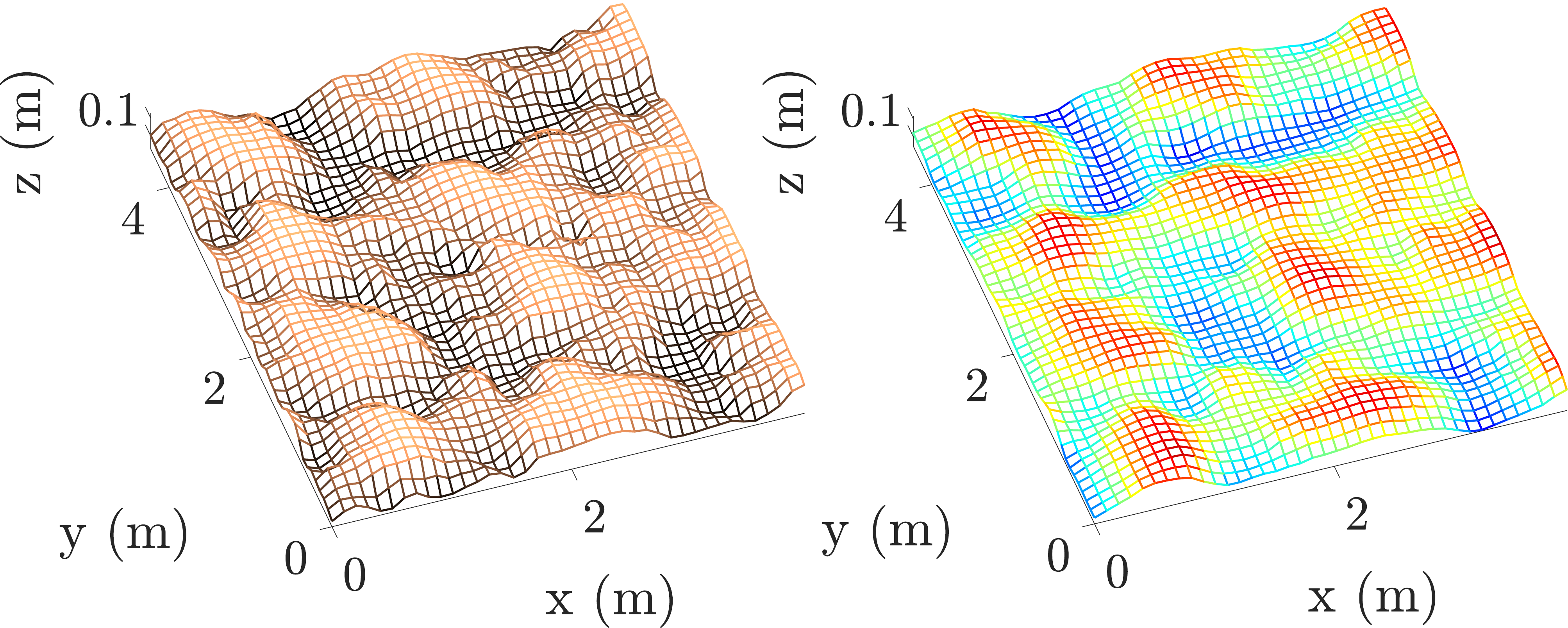}
    \caption{The final terrain estimation results (right) compared with the ground truth (left). The proposed terrain estimation algorithm can effectively estimate rough terrain during walking without prior knowledge.}
    \label{fig: terrain_estimation_experiment}
\end{figure}

\begin{figure}[t]
    \centering
    \includegraphics[width=.9\linewidth]{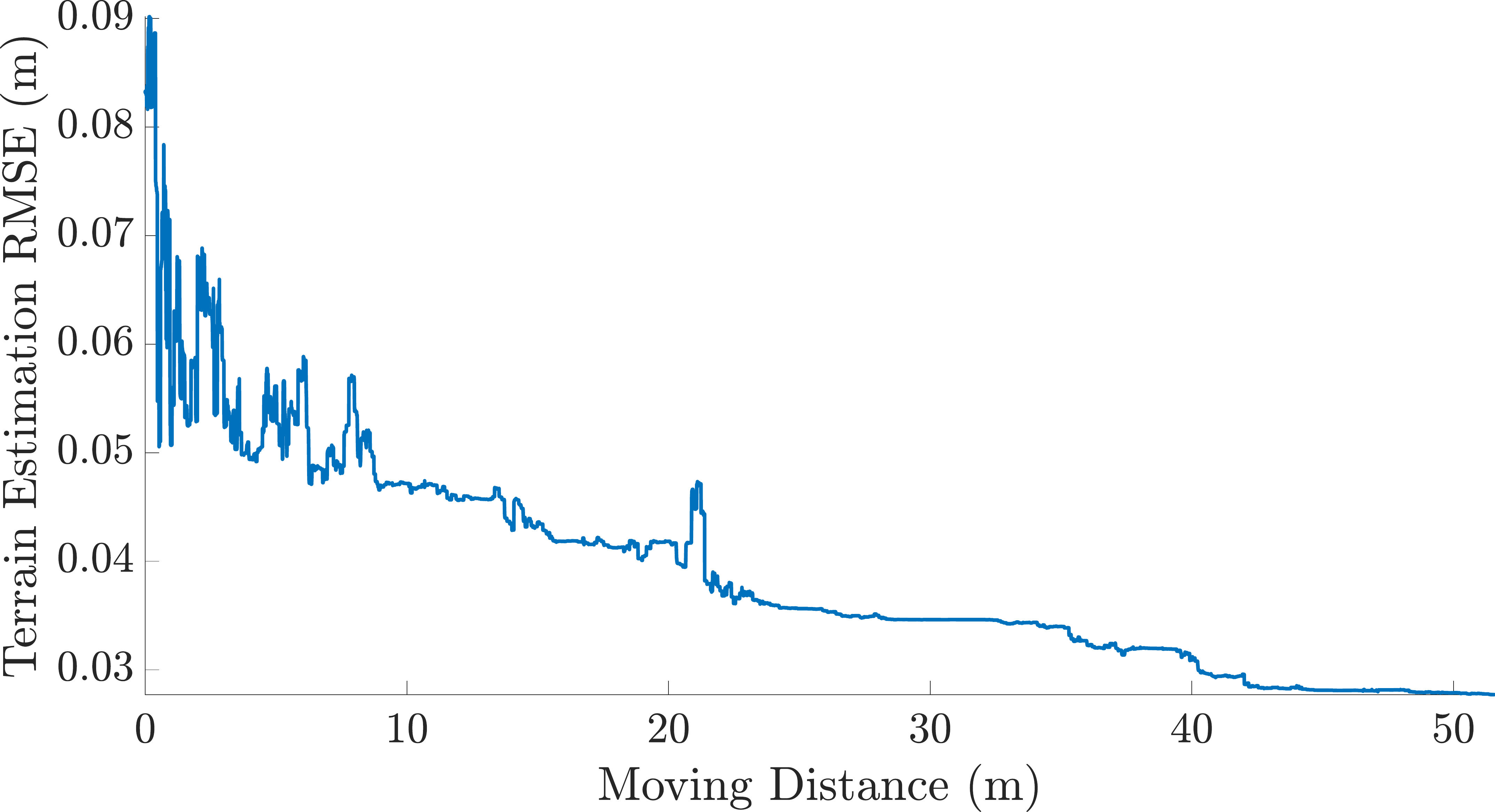}
    \caption{The RMSE of the estimated terrain for the walking area of around 4$\times$4.5m. As the moving distance increases, the overall estimation error effectively converges to an acceptable range. After a 52-meter walk, the final estimate of the terrain has an RMSE of 0.028m.}
    \label{fig: terrain_estimation_experiment_rmse}
\end{figure}

\section{Experimental Evaluation}
\label{sec: experimental_evaluation}


We present 4 simulation experiments to evaluate the proposed perception and control algorithms on challenging terrains. In the experiments, the robot's dynamics are set according to Ghost Robotics Spirit 40 in Quad-SDK \cite{norby2022quad}. All tests are executed in Gazebo 9 with the ODE physics engine on a machine with an Intel Core i7-1165G7 CPU at 2.8 GHz and with 16 GB of RAM. The contact forces for proprioception are obtained with the contact plugin in Gazebo. The tail design, \added{as shown in the right panel of Fig.~\ref{fig: motivation}}, is a long (about twice the body length) and light (\added{a point mass of around 2\% of robot weight attached to a massless rod}) 2DOF tail and attached through roll and pitch motors in series to the top center of the robot, which better approximates decoupled dynamics \cite{libby2016comparative, heim2016designing, saab2018robotic}. Tail and body collisions are modeled by \added{imposing a $\pm$90-degree limit on each joint} in the simulation. To avoid unfair comparison due to the tail supporting the body, the modeling of tail collision with the ground is turned off, but tail joint and body rotation limits work together to avoid the tail colliding with the ground. The simulation uses the full multi-body robot dynamics, while the controller assumes that the leg mass is concentrated at the base of the leg and attached to the body. \added{In the implementation, the contact sensing finite-state machine and gait planner are updated as the leg NMPC iterates, while the terrain estimation is updated at the frequency of sensor sampling. To mitigate the influence caused by state estimation, the robot's state is directly obtained from the ground truth in the simulation.}

\subsection{Terrain Estimation Performance}
\label{sec:terrainresults}

This section evaluates the performance of the proprioceptive terrain estimation algorithm. A simulated robot was manually driven around a patch of rough terrain without any prior knowledge of elevation. The area is about 4$\times$4.5m, and the height standard deviation of the terrain is 50cm, \added{which is} 1.25 times the leg length. Using the proprioceptive terrain estimation proposed in this paper, the robot was able to build an estimate of the local terrain, as illustrated in Fig.~\ref{fig: terrain_estimation_experiment}.

\looseness=-1The root-mean-square error (RMSE) of the elevation estimation over the walking area is illustrated in Fig.~\ref{fig: terrain_estimation_experiment_rmse}, where the error converges to below \added{30cm} after a walk of about 52\added{m}. 
The robot initially only had foothold information right where it started, resulting in large errors and unstable estimation and planning. When \added{traversed} area became larger and the number of recorded footholds increased, the estimation was more stable and could capture the global and local features of the terrain. As illustrated in Fig.~\ref{fig: terrain_estimation_experiment}, the final terrain estimate successfully models peaks and valleys in rough terrain, providing a good reference for future control and planning.

\subsection{Sequential Distributed NMPC \deleted{Tail Controller} Performance}
\label{sec:nmpc_comparison}

\begin{figure}[t]
    \centering
    \includegraphics[width=\linewidth]{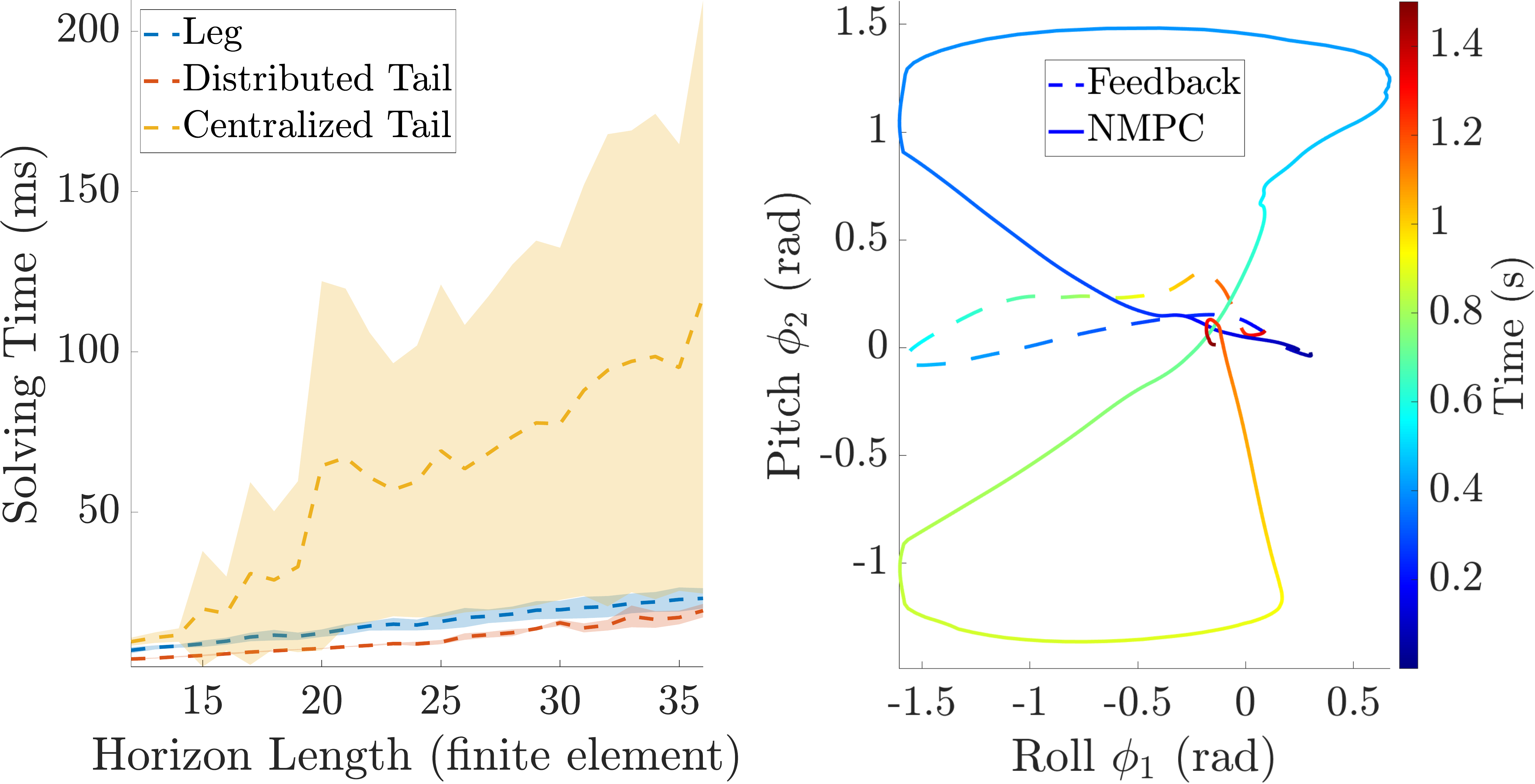}
    \caption{\added{Left:} Solving times for different types of NMPCs at different prediction horizon lengths when missing contacts. The curves show the mean of all trials initialized with each finite element of the gait cycle (12 in total), and the shaded area shows the standard deviation. \added{Right:} The tail swing trajectory of NMPC and feedback controller. The NMPC swung the 2DOF tail in an ``8'' shape to provide prolonged roll maneuvers and maintain pitch stability. It gains small inertia in large pitch positions and returns the roll joint rapidly to accommodate limited angular deflection. The feedback controller, on the other hand, swung only proportional to the current body orientation and angular velocity errors.}
    \label{fig: nmpc_experiment_comparison}
\end{figure}

To demonstrate the advantages of the proposed sequential distributed NMPC for tailed robots, we compare it with a simple feedback controller that swings the tail \deleted{in roll and pitch directions} in proportion to the body orientation and angular velocity as well as a centralized NMPC\footnote{\added{The centralized NMPC shares the same system and constraints as the tail NMPC in~(\ref{eq:mpc}), with the exception that the GRF is not known and is simultaneously determined with the tail torque.}} that solves both leg and tail control simultaneously. The experiment was set at a time when the robot missed contact and needed reactive control to balance itself. Because the leg was not on the ground, the tail had to provide the control authority. More importantly, a missed contact invalidates previous contact schedules and disrupts warmstarting. Therefore, NMPC should solve and update the control commands as soon as it senses a disturbance. 


The efficiency of sequential distributed and centralized NMPC is compared under different prediction horizon lengths. The gait cycle is fixed to 12 finite elements. For each prediction horizon, statistical results are calculated based on an average from 12 trials \added{corresponding to initialization} from each finite element of the gait cycle. The contact miss time within the horizon is set to be half as long as the gait cycle (6 finite elements).  The mean and standard deviation of solving times for leg NMPC, sequential distributed NMPC for tail, and centralized NMPC for leg and tail are illustrated \added{in the left panel of} Fig.~\ref{fig: nmpc_experiment_comparison}. 

The proposed sequential distributed NMPC is more efficient and stable than the centralized NMPC when the contact schedule changes due to missed contact. Its solving time scales with prediction horizon length much better than centralized NMPC and even faster than leg NMPC (since the leg NMPC is a larger optimization problem). The main reason for the improvement is that it can make use of the results of the leg NMPC to warmstart, and the dynamics of the tail itself are continuous so its control should not be affected too much by changes in the contact schedule. Note that because a sequential distributed NMPC separates the control of the legs and tails, its complete control loop requires both NMPCs to be solved, but the two times may not need to be added since they run asynchronously and in parallel. Sequential distributed NMPC brings many benefits -- decoupled tail and leg control allows for faster response to update previous solutions that are invalidated due to missed contacts, more efficient solving allows for longer prediction horizons, and stable solving times reduce deployment uncertainty.


\added{The right panel of} Fig.~\ref{fig: nmpc_experiment_comparison} plots the tail trajectory when the robot fell off a cliff sideways, as \added{in the left panel of} Fig.~\ref{fig: motivation}, for both the sequential distributed NMPC and the feedback controller. The feedback controller performed a single stroke to counteract body tilt in the roll direction. However, its maneuverability is limited due to joint limitations, and the return stroke canceled out most of the positive work of the forward stroke. In contrast, the NMPC made better use of the 2DOF tail dynamics by combining the roll and pitch motions to get two strokes that combine their effect in roll while canceling their effects on pitch. The result is a trajectory similar to a ``figure 8''. Controlling the tail through this nonholonomic behavior improves tail maneuverability under limited rotation angles, and its behavior is consistent with the results of \cite{patel2015conical}. Overall, NMPC provides a more effective, biologically plausible, and physically feasible behavior compared to feedback controllers.

\subsection{Proprioception-Based Gait Planner Performance}
\label{sec:proprioresults}

To evaluate the proposed proprioception-based gait planner on overall performance, we compare it to a vanilla version assuming flat ground at the lowest foot level and nominal contact schedules and footholds. We measure performance by simulating a quadruped walking over an unforeseen elevation change over \added{100} trials\deleted{(N=100)}. To measure the stability of the system under different disturbances, we varied the elevation change from 25 to 60cm, equivalent to about 0.6 to 1.5 times the full length of the leg. The initial position of the robot was varied between trials so that it could reach the edge at any point in the gait cycle. Considering the robot's abduction joint limit is limited to 40.5 degrees, we defined the failure criterion as a maximum roll error higher than 45 degrees as beyond this angle the legs will likely not maintain proper support within the friction cone. \added{All simulations were conducted using a time scale of 0.1$\times$realtime to ensure sufficient time for NMPC to solve for an optimal solution.} For efficiency issues, please refer to Sec.~\ref{sec:nmpc_comparison}.



\begin{figure}[t]
    \centering
    \includegraphics[width=\linewidth]{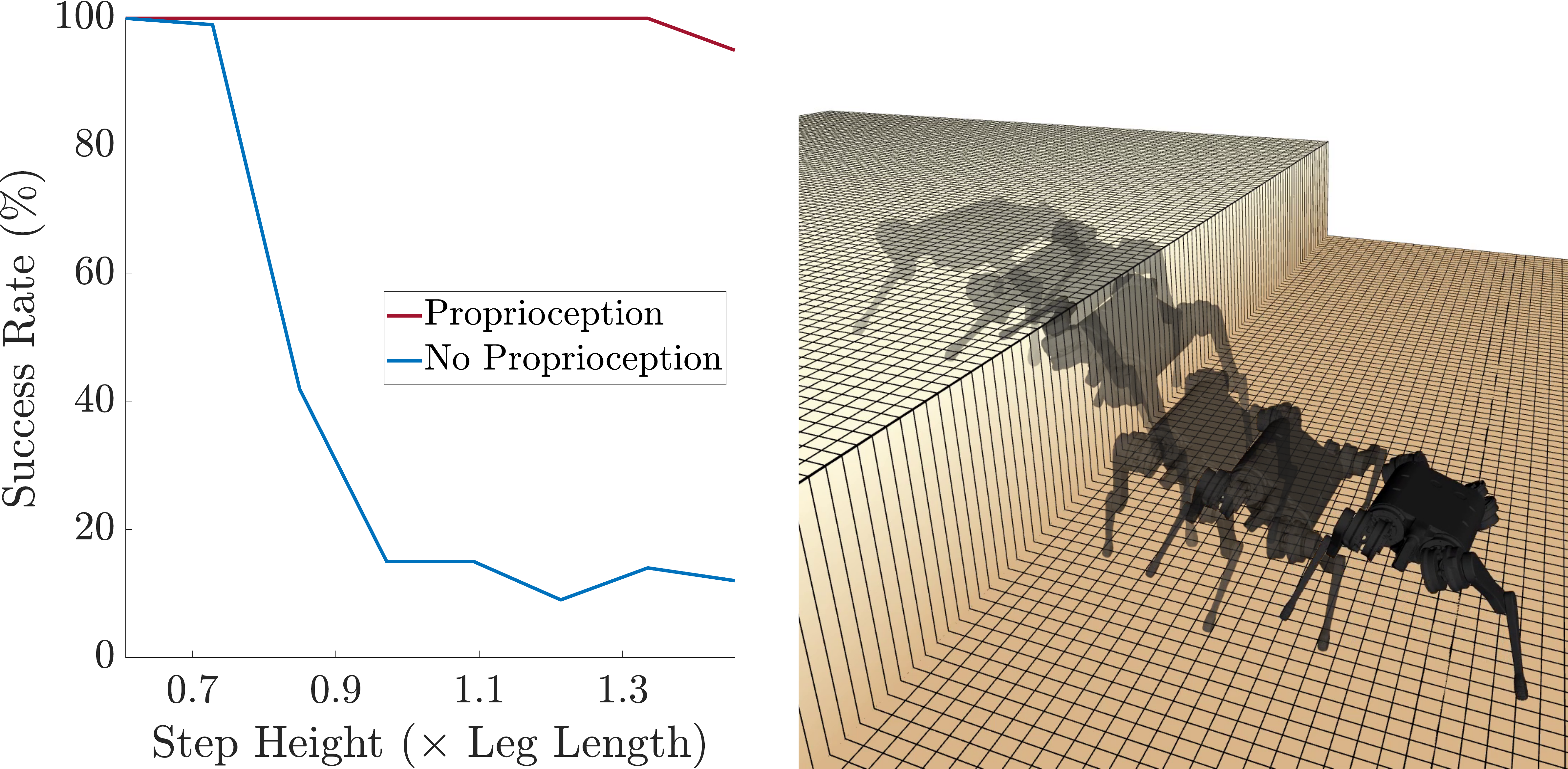}
    \caption{\added{Left:} Batch simulation success rate statistics at different cliff heights with and without proprioception-based gait planner. \added{Right:} Control behavior of the proposed proprioception-based gait planner when the robot walked sideways down an unforeseen cliff. Proprioception recognized the loss of contact, and the gait planner modified the contact schedule, repositioned the legs, and lifted the trapped leg early. The robot landed in an appropriate posture, the gait planner provided supportive gait and control, and then gradually converged back to the clock-based gait.}
    \label{fig: success_rate_proprioception}
\end{figure}

\looseness=-1The success rate statistics are illustrated \added{in the left panel of} Fig.~\ref{fig: success_rate_proprioception}, and an example trial is shown \added{in the right panel of} Fig.~\ref{fig: success_rate_proprioception}. The success rate without the proposed proprioception-based gait planner drops significantly with increasing cliff height, especially when it exceeds the full length of the leg. This is mainly because the contact schedule is increasingly different from the nominal, leaving the basin of attraction for the stability of the controller. In addition, the flat terrain assumption can lead to ineffective foothold and body references, resulting in aggressive or infeasible controls. On the other hand, using a proprioception-based gait planner, the robot can robustly walk over ledges up to 1.5 times the leg length with a success rate of over 95\%. As illustrated \added{in the right panel of} Fig.~\ref{fig: success_rate_proprioception}, the robot successfully identified contact deviations when stepping in the air near the edge of the cliff, and responded by adjusting the contact schedule, repositioning the feet, lifting the trailing legs on top of the cliff early, and balancing the landing. Note that higher cliffs will cause the motors to saturate, so even if the gait planner can maintain orientation, the robot will still fail due to having too much kinetic energy for the motors to absorb.

\subsection{Tailed Robot Control Performance}
\label{sec:tailresults}

In this section, we examine the further performance gains enabled by the tail. The experimental setup and failure conditions are the same as the gait planner analysis, but the robot was initialized with a prior roll error of 0 and 17.5 degrees to compare the success rate of proprioception-based gait planners with and without tails. The initial orientation error is mainly used to simulate the \added{non-ideal} initial conditions of falling caused by uneven terrain in real-world environment tasks. For a legged robot, once it loses contact, its legs cannot provide ground reaction forces to balance the body, so poor posture can easily lead to a failed landing.



\begin{figure}[t]
    \centering
    \includegraphics[width=\linewidth]{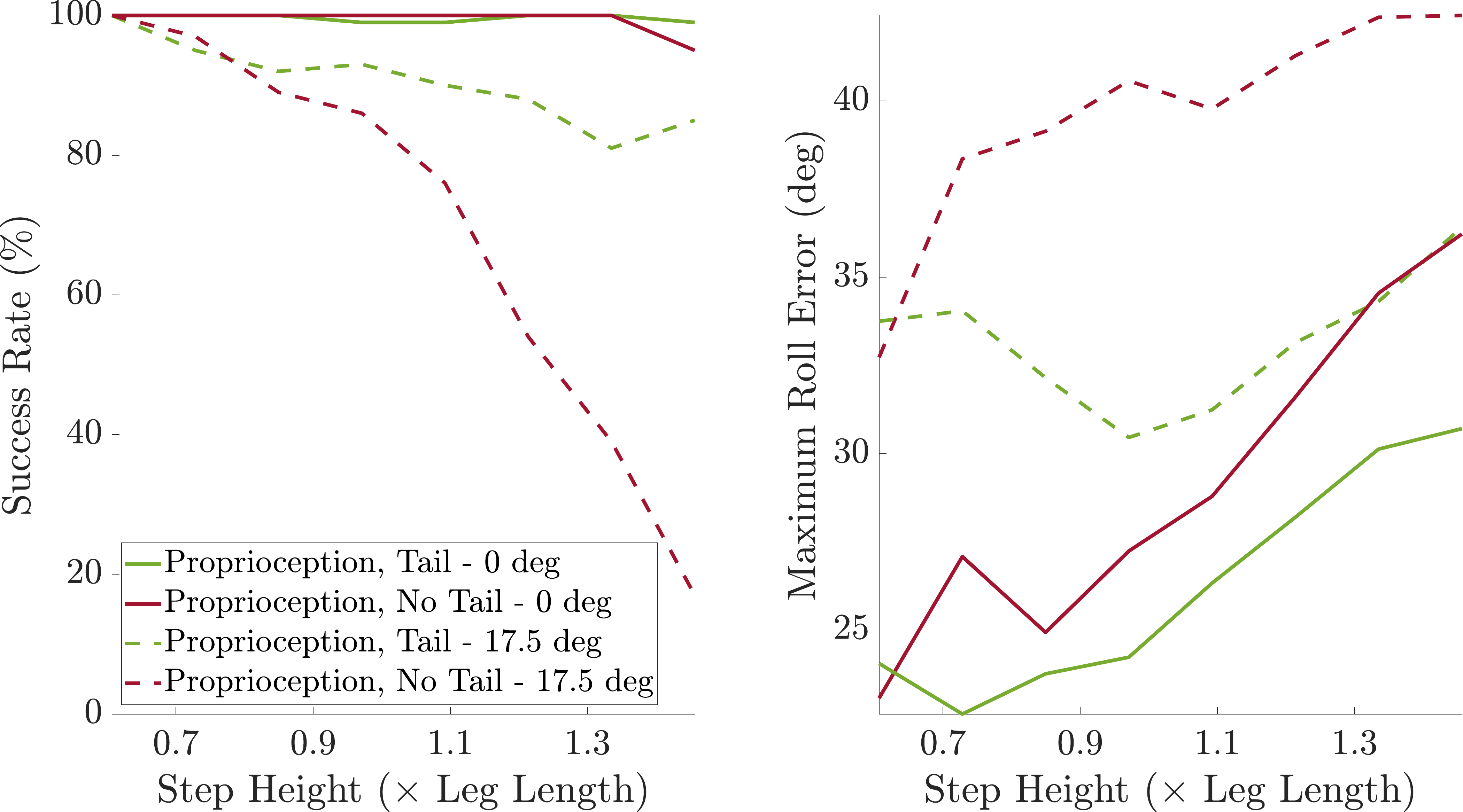}
    \caption{\added{Left:} Batch simulation success rate statistics under different initial conditions and cliff heights with or without a tail. \added{Right:} Average maximum roll error for successful trials at different initial conditions and cliff heights with or without a tail.}
    \label{fig: simulation_comparison_success_rate_tail}
\end{figure}

The success rate statistics are illustrated \added{in the left panel of} Fig.~\ref{fig: simulation_comparison_success_rate_tail}. The success rate without the tail drops significantly with increasing cliff height due to the initial error even with the proposed gait planner. This is because the initial angular disturbance rotated the robot towards the limit of the leg's abduction joint. The poor landing pose violates the support polygon and causes the legs to be over-compressed and unable to cycle. When the tail is included, the success rate increases from 17\% to 85\% for the highest cliff. 

\added{The right panel of} Fig.~\ref{fig: simulation_comparison_success_rate_tail} shows the average maximum roll error for successful trials. 
In the presence of initial roll errors, the tail effectively reduced angular error by up to 25\%.
This will in turn enhance leg control -- a smaller angular error means a better landing posture, and the legs can be positioned over a greater range to balance the body. As illustrated \added{in the left panel of} Fig.~\ref{fig: motivation}, the tail augmented the legs by rejecting angular errors and providing a more kinematically feasible landing. 


\section{Conclusion and Future Work}
\label{sec: conclusion_and_future_work}

This work presents an approach that combines proprioception and tail control to improve extreme terrain traversal by quadrupedal robots. It exploits proprioception to update contact schedules, foothold, and terrain estimation online, improving robustness to unknown elevation changes. Tailed robot control improves performance under angular disturbances by complementing the underactuated legs. These advantages are demonstrated on a simulated quadruped robot that robustly traverses an environment with height variations up to 1.5 times the leg length in a tilted initial state.

\looseness=-1Based on these promising results, one of the main areas of future work is to evaluate the proposed method on hardware. This will place higher demands on state and contact estimation as the control, terrain estimation, and gait planners all highly rely on this information. Additionally, this may require better modeling of the tail to account for its gearbox, as well as finer control constraints to handle tail-body and tail-ground collisions. \added{Finally, it is important to ensure the efficient and feasible solving of NMPC with limited computing resources.} One drawback of this method is that it assumes simple terrain variation. Especially on terrains such as slender ravines or stepping stones, the legs can trip over and never touch the ground. Fortunately in this case, even though the method cannot obtain touchdown information for terrain estimation, the controller will still conservatively lower the body as it learns that the feet have lost contact. This can be further refined by adding events in gait planning, and the experiment can be extended to other types of terrain.

\bibliographystyle{IEEEtran}
\bibliography{ref}

\end{document}